\pdfoutput=1

\documentclass[11pt]{article}

\usepackage{acl}

\usepackage{times}
\usepackage{latexsym}
\usepackage{amsmath}
\usepackage{amsfonts}
\usepackage[T1]{fontenc}
\usepackage[utf8]{inputenc}
\usepackage{microtype}
\usepackage{inconsolata}
\usepackage{graphicx}
\usepackage{graphicx}  
\usepackage{multirow}  
\usepackage{lscape}

\title{ImpliHateVid: A Benchmark Dataset and Two-stage Contrastive Learning Framework for Implicit Hate Speech Detection in Videos}

\author{
 \textbf{Mohammad Zia Ur Rehman\textsuperscript{1}},
 \textbf{Anukriti Bhatnagar\textsuperscript{1}},
 \textbf{Omkar Kabde\textsuperscript{2}},
 \textbf{Shubhi Bansal\textsuperscript{1}},
\\
 \textbf{Nagendra Kumar\textsuperscript{1}}
\\
 \textsuperscript{1}Indian Institute of Technology Indore, Indore, India \\
 \textsuperscript{2}Chaitanya Bharathi Institute of Technology, Telangana, India
\\
\small{
{{phd2101201005@iiti.ac.in}, {mt2302101007@iiti.ac.in}, {ugs22058\_cic.omkar@cbit.org.in}, {phd2001201007@iiti.ac.in}, {nagendra@iiti.ac.in}
    }
 }
 }

\begin{document}

\begin{titlepage}
\begin{center}
\vspace*{10pt}
\textbf{\Large ImpliHateVid: A Benchmark Dataset and Two-stage Contrastive Learning Framework for Implicit Hate Speech Detection in Videos}
\vspace*{20pt}

Mohammad Zia Ur Rehman$^{a}$ (phd2101201005@iiti.ac.in)\\ Anukriti Bhatnagar$^a$ (mt2302101007@iiti.ac.in)\\
Omkar Kabde$^b$ (ugs22058\_cic.omkar@cbit.org.in)\\ Shubhi Bansal$^a$ (phd2001201007@iiti.ac.in)\\ Nagendra Kumar$^a$ (nagendra@iiti.ac.in) \\

\hspace{1pt}

\begin{flushleft}
$^a$ Indian Institute of Technology Indore, Madhya Pradesh India\\

$^b$ Chaitanya Bharathi Institute of Technology, Telangana, India\\

\vspace{2cm}
\normalsize
This is the preprint version of the accepted paper.\\
\textbf{Published in ACL 2025} \\
Published version is available at: https://aclanthology.org/2025.acl-long.842.pdf

\end{flushleft}        
\end{center}
\end{titlepage}

\maketitle

\begin{abstract}

The existing research has primarily focused on text and image-based hate speech detection, video-based approaches remain underexplored. In this work, we introduce a novel dataset, ImpliHateVid, specifically curated for implicit hate speech detection in videos. ImpliHateVid consists of 2,009 videos comprising 509 implicit hate videos, 500 explicit hate videos, and 1,000 non-hate videos, making it one of the first large-scale video datasets dedicated to implicit hate detection. We also propose a novel two-stage contrastive learning framework for hate speech detection in videos. In the first stage, we train modality-specific encoders for audio, text, and image using contrastive loss by concatenating features from the three encoders. In the second stage, we train cross-encoders using contrastive learning to refine multimodal representations. Additionally, we incorporate sentiment, emotion, and caption-based features to enhance implicit hate detection. We evaluate our method on two datasets, ImpliHateVid for implicit hate speech detection and another dataset for general hate speech detection in videos, HateMM dataset, demonstrating the effectiveness of the proposed multimodal contrastive learning for hateful content detection in videos and the significance of our dataset. The code and dataset will be made available on the GitHub repository\footnote{https://github.com/videohatespeech/Implicit\_Video\_Hate}. 

\end{abstract}

\section{Introduction}

With approximately 66\% of the world’s population having access to the internet\footnote{https://www.itu.int/en/ITU-D/Statistics/Pages/facts/default.aspx}, online communication has become an integral part of daily life. However, the widespread accessibility of digital platforms has also facilitated the rapid dissemination of hate speech. Despite efforts by online platforms to regulate such content through AI-based detection, manual review, and community reporting \cite{RegulatingOnlineHateSpeechthroughthePrismofHumanRightsLawThePotentialofLocalisedContentModeration}, hateful content remains a persistent challenge due to the vast amount of data generated every day \cite{ibanez2021audio, das2023hatemm, wu2020detection}.

Hate speech is defined as public speech that expresses hate or encourages violence towards a person or group based on race, religion, sex, or caste\footnote{https://dictionary.cambridge.org/us/dictionary/english/hate-speech}. It manifests in multiple forms, including texts, images, memes, gestures, and symbols, both online and offline\footnote{https://www.un.org/en/hate-speech/understanding-hate-speech/what-is-hate-speech}. Most existing research on hate speech detection has focused on textual content, such as tweets and comments \cite{fortuna2018survey, schmidt2017survey}, or image-based hate speech, particularly in memes \cite{cao2023pro, cao2023prompting, sharma2022detecting, nayak2022multimodal, hee2023decoding}. While some studies have explored hate detection in videos \cite{alcantara2020offensive, das2023hatemm, wang2024multihateclip}, implicit hate speech detection in videos remains an underexplored area. To the best of our knowledge, we are the first to work in implicit hate speech detection in videos.

Implicit hate speech is defined as expressions that communicate discriminatory or prejudiced views indirectly, often through coded language, implied meanings, or contextual cues \cite{elsherief2021latent}. Unlike overt hate speech, it subtly evades detection by adhering to platform guidelines and may appear innocuous superficially, yet still perpetuates harm or offense.
 Given the dominance of video content in digital communication, there is a strong need to develop specialized hate detection mechanisms for videos.

\begin{figure}
    \centering
    \includegraphics[width=1\linewidth]{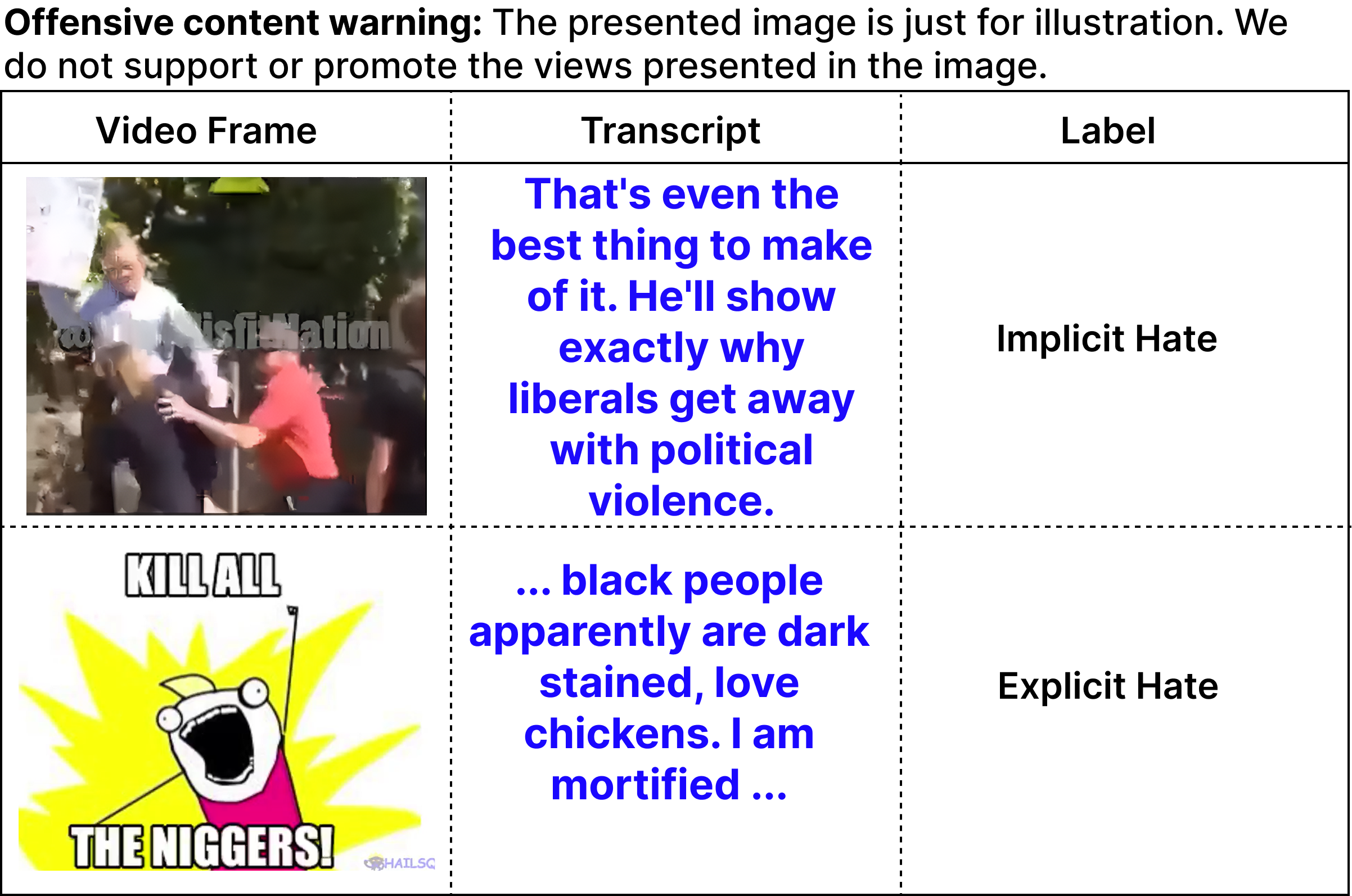}
    \caption{Implicit and Explicit Hate Videos}
    \label{fig:example}
\end{figure}

\autoref{fig:example} illustrates examples of both implicit and explicit hate speech in video content. In the implicit hate example, the hateful intent is conveyed through the underlying context and the association of political ideology with aggression. Such indirect expressions make implicit hate speech particularly challenging to detect, as they require an understanding of both textual and visual cues in the video. Conversely, the explicit hate speech example features a cartoonish character who appears to be shouting a racial slur directed at Black individuals.

This distinction underscores the necessity of advanced multimodal methods for effective implicit hate speech detection in videos.

In this work, we address the critical challenge of detecting hateful content in videos by proposing a novel method with a particular focus on implicit hate speech, which is often subtle and context-dependent. Our two-stage contrastive learning method first extracts robust, modality-specific features from text, images, and audio, and then aligns them in a shared embedding space using cross-modal encoders with projection heads and supervised contrastive loss. This approach captures subtle cues of implicit hate speech that traditional unimodal or simple fusion methods often miss. By leveraging complementary multimodal information, our framework provides a richer, more nuanced representation for hate speech detection in videos.
Our key contributions are as follows:  

\begin{itemize}  
    \item We introduce a new dataset specifically curated for implicit hate speech detection in videos. The dataset consists of 2,009 videos and provides a valuable benchmark for future research on multimodal hate speech detection.  

    \item  We propose a two-stage contrastive learning approach to effectively model multimodal hateful content in videos. In the first stage, we train three modality-specific encoders (audio, text, and image) using contrastive loss, computed over concatenated feature representations. In the second stage, we train a cross-encoder using contrastive learning to refine multimodal representations further.  

    \item  We evaluate our approach on both our newly curated dataset and the publicly available HateMM dataset. The results demonstrate the effectiveness of our proposed multimodal contrastive learning framework in detecting hateful content in videos, particularly implicit hate speech.  

\end{itemize}

\section{Related Works}

\subsection{Implicit Hate Speech Detection}

Recent studies on implicit hate speech detection have primarily focused on text-based approaches. For instance, \cite{elsherief2021latent} and \cite{kim2022generalizable} have advanced the field using linguistic analysis to capture subtle hate cues, while \cite{ocampo2023playing} and \cite{guo2023implicit} further enhanced detection by integrating emotion, ambiguity, and multi-feature fusion techniques. Additionally, efforts such as ToxiGen \cite{hartvigsen2022toxigen} have leveraged machine-generated data to improve model robustness. Despite these advances, the majority of existing work relies solely on textual information, neglecting the rich, multimodal context inherent in video content. In contrast, our study is the first to explore implicit hate speech detection in videos, incorporating not only text but also visual and audio modalities to capture a more comprehensive spectrum of hateful content.

\subsection{Hate Speech Detection in Videos}
Hate speech detection in videos is an emerging field, yet many existing works focus only on explicit hate. For example, HateMM \cite{das2023hatemm} introduced a dataset of 1,083 English BitChute videos aimed at binary hate classification only. Similarly, MultiHateClip \cite{wang2024multihateclip} provides a multilingual dataset of 2,000 YouTube and Bilibili videos with fine-grained labels in English and Chinese, but these labels predominantly target explicit hate content. Additionally, studies by Alcântara et al. \cite{alcantara2020offensive} and Wu et al. \cite{wu2020detection} have advanced video-based hate speech detection; however, they too tend to overlook the subtleties of implicit hate. This gap underscores the need for comprehensive approaches that not only detect overt hateful content in videos but also capture the nuanced and often overlooked manifestations of implicit hate speech.

\begin{table*}[]
    \centering
    \resizebox{1.5\columnwidth}{!}{
    \begin{tabular}{|l|l|l|l|}
    \hline
        \textbf{Properties} & \textbf{Non Hate} & \textbf{Implicit Hate} & \textbf{Explicit Hate}  \\
        \hline
        \textbf{Video count} & 1,000 (49.78\%) & 509 (25.36\%) & 500 (24.89\%)\\
        \textbf{Total length} & 39 hours 26 mins 42 secs & 18 hours 7 mins 51 secs & 28 hours 58 mins 25 secs\\
        \textbf{Mean video length} & 2 mins 22 secs & 2 mins 8 secs & 2 mins 38 secs\\
        \textbf{Total number of frames} & 1,42,002 & 65,271 & 79,105\\
        \textbf{Mean number of frames} & 142.002 & 128.23 & 158.21\\
        \textbf{Mean number of words} & 175.404 & 85.166 & 80.326\\
        \hline
    \end{tabular}}
    \caption{Dataset Statistics}
    \label{tab:Dataset Statistics}
\end{table*}

\section{Data Collection and Annotation}
\subsection{Platforms for Video Collection}
For data collection we primarily used BitChute\footnote{https://www.bitchute.com/}, a social video-hosting platform launched in 2017 with minimal content moderation. It has grown in prominence as an alternative to YouTube, hosting a significant amount of hateful content banned from mainstream platforms. In addition to BitChute, we also collected videos from Odysee\footnote{https://odysee.com/}, another alternative video-sharing platform.

\subsection{Dataset Statistics}

Our dataset consisted of 2,009 videos comprising 86.5 hours of English multimodal content collected from BitChute and Odyssey. Among these, 1,000 videos were classified as non-hate, while the remaining 1,009 videos contained hateful content. The hate videos were further categorized into implicit hate and explicit hate, with 509 and 500 videos, respectively, to capture different degrees of hateful expressions. The details of the dataset can be seen in \autoref{tab:Dataset Statistics}.
The dataset is balanced with almost 50\% hate and non-hate content by count. Duration-wise, we have ~47 hours of hate content compared to ~39.5 hours of non-hate content. Within the hate category, the number of samples of implicit and explicit hate is also roughly balanced. On average, all the videos are approximately 2 minutes long with approximately 143 frames captured per video. Non-hate videos have a little more than double the number of words in transcripts than those in implicit and explicit hate videos.

\subsection{Annotation Guidelines}
The following labeling scheme provided the main framework for annotators, while a codebook ensured consistency in label interpretation as inspired by \cite{das2023hatemm, wang2024multihateclip, salles2025hatebrxplain}. Developed using YouTube’s hate speech policy, the codebook contains detailed annotation guidelines. A video is considered hateful if:

``It promotes discrimination, disparages, or humiliates an individual or group based on characteristics such as race, ethnicity, nationality, religion, disability, age, veteran status, sexual orientation, or gender identity.''

Moreover, the annotators were guided to pinpoint particular segments of a hate video (i.e., frame spans) they deemed hateful and indicate the specific communities targeted by the content.

\subsection{Annotation Process}
\textbf{Annotator Training}\\
The annotation process was supervised by one Professor and one PhD student with expertise in analyzing harmful content on social media, while the actual annotations were carried out by one postgraduate and three undergraduate students who were novice annotators. All annotators were computer science majors and participated voluntarily with full consent. As a token of appreciation, they were rewarded with free access to A100 GPU for 150 hours.

The annotators were trained by creating an initial gold-standard dataset. The expert annotators labeled 50 videos, comprising 30 hate videos and 20 non-hate videos which were then provided to the annotators for labeling based on the annotation codebook. After completing their annotations, we reviewed and discussed the incorrect cases with them to refine their understanding and improve their annotation accuracy.

\hspace{-5 mm}
\textbf{Annotation in Batch Mode}\\
Following the initial training, we adopted a batch-mode approach, releasing a set of 50 videos per week for annotation. Given the potential negative psychological effects of annotating hate content \cite{ybarra2006examining}, we advised annotators to take a minimum 10-15 minute break after labeling each video. Additionally, we imposed a strict limit of no more than 20 videos per day to prevent cognitive overload. To ensure the well-being of the annotators, we also conducted regular check-in meetings to monitor any potential adverse effects on their mental health.

\section{Methodology}
\begin{figure*}
    \centering
    \includegraphics[width=1\linewidth]{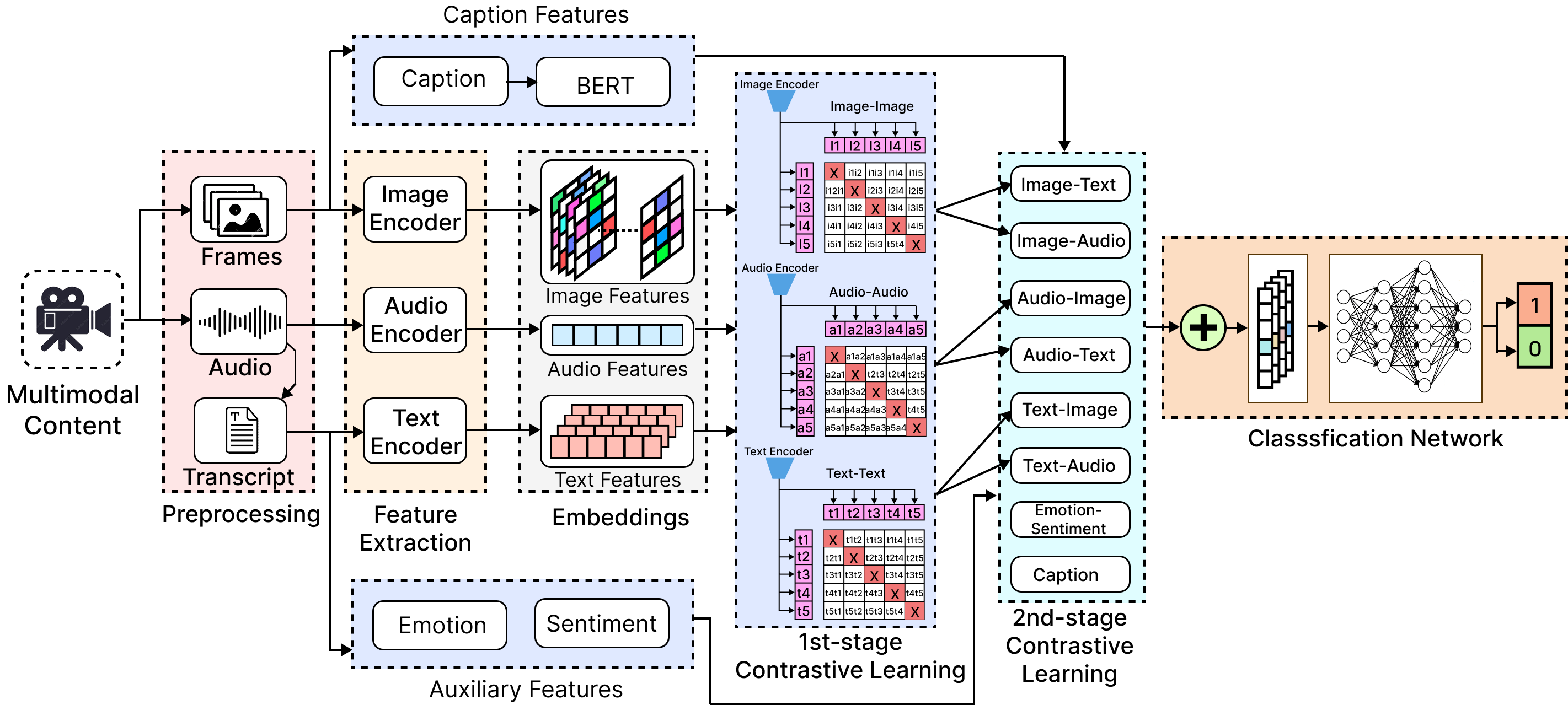}
    \caption{The proposed method first extracts modality-specific features from text, images, and audio using dedicated encoders.  It then aligns these features in a shared embedding space through cross-modal encoders with projection heads and supervised contrastive loss. Additionally, emotion, sentiment and caption features are also extracted.}
    \label{fig:method}
\end{figure*}

We propose a two-stage contrastive learning multimodal framework for detecting hateful content in videos as shown in \autoref{fig:method}. Our approach comprises three key steps: (i) preprocessing to extract audio, text, and visual data; (ii) feature extraction using modality-specific encoders; and (iii) contrastive learning to align representations, culminating in a multimodal classifier.

\subsection{Preprocessing}
Videos are converted to WAV audio using FFmpeg and transcribed via speech-to-text conversion. For the visual modality, 100 frames are uniformly sampled (with padding for videos having fewer frames) using VideoCapture, ensuring consistent input dimensions across samples.

\subsection{Feature Extraction}
We extract features from image, text, and audio using a video-based contrastive learning method, ImageBind \cite{girdhar2023imagebind}, that maps raw inputs into a shared 1024-dimensional space. 
We denote the extracted features with \(f_I\), \(f_T\), and \(f_A\) for the image, text and audio, respectively.

In addition to extracting the 1,024-dimensional audio, text, and video features, we extract complementary features to enrich our multimodal representation. For a text transcription \(x_T\), we compute emotion features \(e = \text{NRCLex}(x_T)\) (with \(e \in \mathbb{R}^{d_e}\)) and a sentiment score \(s = \text{Vader}(x_T)\) (\(s \in \mathbb{R}\)); these are concatenated into a joint representation \(f_{ES} = [e, s] \in \mathbb{R}^{d_e+1}\). Similarly, for an image \(x_I\), we generate a caption \(c = \text{caption\_gen}(x_I)\) and extract caption features \(f_C = \text{BERT}(c)\) where \(f_C \in \mathbb{R}^{d_c}\). Captions are generated through OFA model \cite{wang2022ofa}. Together, these features complement the primary modality representations to provide a comprehensive view of the content.

\subsection{Two-stage Contrastive Learning}

\subsubsection{Stage 1: Modality-specific Encoder Training}
To jointly optimize the image, text, and audio encoders from the first stage, denoted by \(\text{encoder}_{II}\), \(\text{encoder}_{TT}\), and \(\text{encoder}_{AA}\), we merge their outputs via a projection head. Let the encoder outputs be defined as
\(f_{II}\), \(f_{TT}\), and \(f_{AA}\) for the image, text, and audio modalities, respectively, from the first stage encoders. These features, each of dimension \(d\) (e.g., 1,024), are concatenated to form a joint representation:
\begin{equation}
    f_{ITA} = \operatorname{Concat}(f_{II}, f_{TT}, f_{AA}) \in \mathbb{R}^{3d}
\end{equation}

A projection head \(P: \mathbb{R}^{3d} \rightarrow \mathbb{R}^{d'}\) is then applied to map the concatenated features into a shared embedding space:
\begin{equation}
    z_{ITA} = P(f_{ITA}) \in \mathbb{R}^{d'}
\end{equation}

The merged encoder (denoted as ITA) is optimized using a supervised contrastive loss. Specifically, for a batch of \(N\) samples, the loss is defined as:
\begin{equation}
    \small
    \mathcal{L}_{\text{sup}} = -\frac{1}{N} \sum_{i=1}^N 
    \log \frac{\displaystyle\sum_{j \in \mathcal{P}(i)} \exp\left(\frac{\text{sim}(z_{ITA}^i, z_{ITA}^j)}{\tau}\right)}
    {\displaystyle\sum_{k \in \mathcal{N}(i)} \exp\left(\frac{\text{sim}(z_{ITA}^i, z_{ITA}^k)}{\tau}\right) + \epsilon}
\end{equation}
where:
 \(\text{sim}(z_{ITA}^i, z_{ITA}^j)\) is the cosine similarity between embeddings, \(\tau\) is the temperature parameter, \(\epsilon\) is a small constant for numerical stability, \(\mathcal{P}(i)\), and \(\mathcal{N}(i)\) denote the sets of indices corresponding to positive (same class) and negative (different classes) pairs for sample \(i\), respectively. 
 In this way, the individual encoders \(\text{encoder}_{II}\), \(\text{encoder}_{TT}\), and \(\text{encoder}_{AA}\) are optimized jointly using the merged ITA representation, thereby aligning multimodal features in a unified embedding space. Moreover, specialized encoders for emotion-sentiment (\(ES\)) and image captions (\(CP\)) are trained in a similar manner.

\subsubsection{Stage 2: Cross-Modal Encoder Training}

In this stage, we align representations across modalities by training cross-modal encoders that merge outputs from modality-specific encoders via a projection head. We can understand this through an example of image\_text cross encoders.
The features from 1st-stage encoders are first processed by cross encoders \(\text{encoder}_{IT}\) and \(\text{encoder}_{TI}\) to produce refined features \(f_{IT}\) and \(f_{TI}\):
\begin{equation}
f_{IT} = \text{encoder}_{IT}(f_{II})    
\end{equation}
\begin{equation}
    f_{TI} = \text{encoder}_{TI}(f_{TT})
\end{equation}

Next, a projection head \(P(\cdot)\), implemented as a dense layer with ReLU activation, maps these to a shared embedding space:
\begin{equation}
    z_{IT} = P(f_{IT})
\end{equation}
\begin{equation}
    z_{TI} = P(f_{TI})
\end{equation}

The embeddings are concatenated to form the cross-modal representation:
\begin{equation}
    z_{\text{cross}} = \operatorname{Concat}(z_{IT}, z_{TI})
\end{equation}

Subsequently, the cross-modal encoder is optimized using a supervised contrastive loss. In addition to the image\_text encoders, we similarly train cross encoders for image\_audio (\(IA\) and \(AI\)) and text\_audio (\(TA\) and \(AT\)) pairs. 

The overall training objective for our system combines the losses from the first stage (modality-specific encoders), the cross-modal encoders, and the specialized \(ES\) and \(CP\) encoders:
\begin{equation}
\mathcal{L}_{\text{total}} = \mathcal{L}_{\text{stage 1}} + \mathcal{L}_{\text{stage 2}} + \mathcal{L}_{\text{sup}}^{ES} + \mathcal{L}_{\text{sup}}^{CP}
\end{equation}

By jointly optimizing these components, the framework effectively aligns and integrates multimodal features from image, text, and audio. This unified representation enhances the model's ability to capture complementary information, thereby improving performance in tasks such as detecting hateful content in videos.

\begin{table*}
  \centering
  \resizebox{0.7\textwidth}{!}{  
  \begin{tabular}{|l|l|cccc|cccc|}
    \hline
    \multirow{2}{*}{\textbf{Modality}} & \multirow{2}{*}{\textbf{Method}} & 
    \multicolumn{4}{|c|}{\textbf{ImpliHateVid Dataset}} &  \multicolumn{4}{|c|}{\textbf{HateMM Dataset}}\\
    & & Acc & F1 & Prec & Rec & Acc &  F1 & Prec & Rec\\
    \hline
    \multirow{3}{*}{\textbf{Text}}
    & BERT & 0.6907 & 0.6884 & 0.6907 & 0.6907  & 0.7350 & 0.6640 & 0.6750 & 0.6670\\
    & GPT-4o & 0.5312 &  0.1132  & \textbf{1.0000} & 0.0600 & 0.5652 & 0.1964 & 0.3793 & 0.1325\\
    & Llama 3.1-8b & 0.5312 &  0.2034  & 0.6667  & 0.1200 & 0.5459 & 0.2540 & 0.3721 & 0.1928  \\
    \hline
    \multirow{2}{*}{\textbf{Image}} 
    & ViT & 0.7655  &  0.7684  & 0.7658  & 0.7656  & 0.7480  & 0.6720 & 0.6950 & 0.6560 \\
    & ViViT & 0.4912  & 0.5255 & 0.4912 & 0.4914 & 0.5293 & 0.5176 & 0.5172 & 0.5182\\
    \hline
    \multirow{2}{*}{\textbf{Audio}}
    & MFCC & 0.4987 &  0.6655 & 0.2493 & 0.5000 & 0.6750 & 0.6220 & 0.5930 & 0.6790\\
    & Wav2Vec2 & 0.7531 & 0.7724 & 0.7610 & 0.7533 & 0.5810 & 0.5810 & 0.5270 & 0.5160\\
    \hline
    \multirow{2}{*}{\textbf{Video}} 
    & GPT-4 & 0.4988 & 0.6656 & 0.4988 & \textbf{1.0000} & 0.4010  & 0.5724 & 0.4010 & \textbf{1.0000}\\
    & LlamaVL & 0.4010 & 0.5724 & 0.4010 & \textbf{1.0000} & 0.3800 & 0.5300 & 0.3700 & 0.9500\\
    \hline
    \multirow{6}{*}{\textbf{Multimodal}} 
    & DeepCNN & 0.7623 & 0.7800 & 0.7481 & 0.7933 & 0.5622 & 0.3065 & 0.4565 & 0.2307\\
    & CMHFM & 0.7922 & 0.7921 & 0.7860 & 0.7980 & 0.6057 & 0.5629 & 0.6184 & 0.6184\\
    & CSID & 0.8150 & 0.8154 & 0.8082 & 0.8233 & 0.7320 & 0.7140 & 0.7200 & 0.7230\\
    & MCMF & 0.8224 & 0.8220 & 0.8200 & 0.8240 & 0.5769 & 0.0435 & 0.5000 & 0.2422\\
    & MulT & 0.8352 & 0.8352 & 0.8320 & 0.8380 & 0.6571 & 0.5212 & 0.4318 & 0.6571\\
    & \textbf{Proposed Method} & \textbf{0.8753} & \textbf{0.8773} & 0.8796 & 0.8752 & \textbf{0.9758} & \textbf{0.9758} & \textbf{0.9745} & 0.9710\\
    \hline
  \end{tabular}
  }
  \caption{\label{tab:performance-comparison}
    Effectiveness Comparison for Binary Classification across Different Methods and Datasets
  }
\end{table*}

\begin{table*}
  \centering
  \resizebox{\textwidth}{!}{
  \begin{tabular}{|l|l|cccc|cccc|cccc|c|}
    \hline
    \multirow{2}{*}{\textbf{Modality}} & \multirow{2}{*}{\textbf{Method}} & \multicolumn{4}{|c|}{\textbf{Non Hate Videos}} &  \multicolumn{4}{|c|}{\textbf{Implicit Hate Videos}} & \multicolumn{4}{|c|}{\textbf{Explicit Hate Videos}} & \textbf{Overall}\\
    & & Acc & F1 & Prec & Rec & Acc  & F1 & Prec & Rec & Acc  & F1 & Prec & Rec & Macro-F1 \\
    \hline
    \multirow{3}{*}{\textbf{Text}} 
    & BERT & 0.7195 & 0.7192 & 0.7122 & 0.7264 & 0.7107  & 0.2927 & 0.4138 & 0.2264 & 0.7207 & 0.5172 & 0.4348 & 0.6383 & 0.5907 \\
    & GPT-4o & 0.5362 & 0.6804 & 0.5197 & \textbf{0.9851} & 0.7282 & 0.0684 & 0.3636 & 0.0377 & \textbf{0.7880} & 0.1748 & \textbf{1.0000} & 0.0957 & 0.3078 \\
    & Llama 3.1-8b & 0.5771 & 0.5066 & 0.4514  & 0.5771  & 0.0189 & 0.0231 & 0.0299 & 0.0189 & 0.4043 & 0.4444 & 0.4935 & 0.4043 & 0.3247 \\
    \hline
    \multirow{2}{*}{\textbf{Image}} 
    & ViT & 0.7805 & 0.7854 & 0.7703 & 0.8010 & 0.7307  & 0.4906 & 0.4906 & 0.4906 & 0.7706 & 0.4889 & 0.5116 & 0.4681 & 0.5883 \\
    & ViViT & 0.5012 & 0.5745 & 0.5019 & 0.6716 & 0.6559 & 0.1786 & 0.2419 & 0.1415 & 0.6708 & 0.1951 & 0.2286 & 0.1702 & 0.3161 \\
    \hline
    \multirow{2}{*}{\textbf{Audio}} 
    & MFCC & 0.5262 & 0.6769 & 0.5142 & 0.9900 & \textbf{0.7357} & 0.2563 & 0.3180 & 0.2146 & 0.7506 & 0.0741 & 0.2857 & 0.0426 & 0.2503 \\
    & Wav2Vec2 & 0.7781 & 0.7963 & 0.7357 & 0.8657 & \textbf{0.7357} & 0.3117 & 0.5000 & 0.2264 & 0.7930 & \textbf{0.6066} & 0.5470 & \textbf{0.6809} & 0.5716 \\
    \hline
    \multirow{2}{*}{\textbf{Video}} 
    & GPT-4 & 0.4938 & 0.6381 & 0.4972 & 0.8905 & 0.7082 & 0.0488 & 0.1765 & 0.0283 & 0.7556 & 0.1695 & 0.4167 & 0.1064 & 0.2855 \\
    & Llama-VL & 0.4250 & 0.4800 & 0.4000 & 0.7800 & 0.2500 & 0.0250 & 0.1000 & 0.0150 & 0.3800 & 0.1500 & 0.3800 & 0.1000 & 0.2180 \\
    \hline
    \multirow{6}{*}{\textbf{Multimodal}} 
    & DeepCNN & 0.7623 & 0.7512 & 0.7634 & 0.7398 & 0.6785 & 0.6345 & 0.6189 & 0.6512 & 0.6612 & 0.5803 & 0.5692 & 0.5917 & 0.6587 \\
    & CMHFM & 0.7645 & 0.7534 & 0.7701 & 0.7405 & 0.6802 & 0.6372 & 0.6241 & 0.6509 & 0.6634 & 0.5814 & 0.5723 & 0.5922 & 0.6604 \\
    & CSID & 0.7658 & 0.7556 & 0.7714 & 0.7437 & 0.6814 & 0.6394 & 0.6273 & 0.6534 & 0.6649 & 0.5834 & 0.5745 & 0.5938 & 0.6621 \\
    & MCMF & 0.7661 & 0.7568 & 0.7735 & 0.7452 & 0.6819 & 0.6401 & 0.6289 & 0.6541 & 0.6652 & 0.5845 & 0.5751 & 0.5942 & 0.6625 \\
    & MulT & 0.7667 & 0.7574 & 0.7741 & 0.7459 & 0.6823 & 0.6408 & 0.6296 & 0.6548 & 0.6658 & 0.5849 & 0.5756 & 0.5948  & 0.6627 \\
    & \textbf{Proposed Method} &  \textbf{0.8955} & \textbf{0.8448} & \textbf{0.8000} & 0.8955 & 0.6698 & \textbf{0.6605} & \textbf{0.6513} & \textbf{0.6698} & 0.4894 & 0.5702 & 0.6866 & 0.4894 & \textbf{0.6918} \\
    \hline
  \end{tabular}
  }
  \caption{\label{tab:multiclass-performance} Effectiveness Comparison for Multiclass Classification across Different Methods on ImpliHateVid Dataset}
\end{table*}

\subsection{Multimodal Classification}
The learned representations \(f_{IT}\), \(f_{IA}\), \(f_{TI}\), \(f_{TA}\), \(f_{AI}\), \(f_{AT}\), \(f_{ES}\), and \(f_{CP}\) are concatenated to form a unified feature \(F\). This vector is passed through dense layers with ReLU activations and dropout regularization to produce the final prediction:
\begin{equation}
\begin{aligned}
    y = &\operatorname{softmax} \Bigl( W_4\,\text{Dropout} \bigl( \sigma(W_3\,\text{Dropout} (\sigma(W_2 \\&\,\text{Dropout} (\sigma(W_1 F + b_1)) + b_2)) 
    + b_3) \bigr) + b_4 \Bigr)
\end{aligned}
\end{equation}

where \(y \in \mathbb{R}^C\) represents the predicted probability distribution over \(C\) classes.

This streamlined framework effectively integrates multimodal data and contrastive learning to improve the detection of hateful content in videos.

\section{Results}
\subsection{Datasets}
In addition to the proposed ImpliHateVid dataset, we used HateMM \cite{das2023hatemm}, another multimodal dataset to evaluate the performance of our model. HateMM consists of 1,083 videos in total comprising 43.26 hours of multimodal content in English collected from BitChute and Odysee platforms. Out of the 1,083 videos, 431 videos have been labeled as hate videos and 652 videos have been labeled as non-hate videos. However, we used only 1,035 videos out of 1,083 for our study.

\subsection{Experimental Setup}
We split both datasets into train, validation, and test sets containing the total videos, respectively. The HateMM dataset had 662, 166, and 210 samples in training, validation, and test sets, respectively, while our dataset, ImpliHateVid, had 1,283 samples in the training set, 325 samples in the validation set, and 401 samples in the test set. 
 We experimented with several combinations of hyperparameters. 32 and 64 were used as batch sizes. Our learning rate was in the range {1e-3,1e-4,1e-5}, and we trained our model for 30, 50, 75, and 100 epochs. Adam optimizer \cite{diederik2014adam} was used. We used Accuracy (Acc), Precision (Prec), Recall (Rec), F1-score (F1), and Macro-F1 metrics to evaluate the performance.

\subsection{Compared Methods}
We have compared our proposed method with several unimodal and multimodal methods to demonstrate the effectiveness of our approach. For the textual modality, we evaluated the performance of \textbf{BERT} \cite{kenton2019bert} alongside large language models (LLMs), including \textbf{GPT-4o} \cite{radford2018improving} and \textbf{Llama 3.1-8b} \cite{touvron2023llama}. For the vision modality, we utilized \textbf{ViT} \cite{dosovitskiy2020image} and \textbf{ViViT} \cite{arnab2021vivit}, while for the audio modality, we employed \textbf{MFCC} \cite{jung2021efficiently} and \textbf{Wav2Vec2} \cite{baevski2020wav2vec}. To assess video performance holistically, we also incorporated vision-language models such as \textbf{GPT-4}\footnote{https://cdn.openai.com/papers/GPTV\_System\_Card.pdf} and \textbf{Llama-VL} \cite{zhang2023video}. Features extracted from these models (excluding GPT and Llama) were fed into a feedforward neural network for classification, comprising four dense layers with 512, 256, 128, and 64 neurons, respectively, and a dropout rate of 0.3. GPT and Llama models were used using APIs with zero-shot prompting. Additionally, we compared our proposed method against other multimodal approaches, including \textbf{DeepCNN} \cite{dixit2024deep} and \textbf{CMHFM} \cite{wang2024cross}. The model proposed by \cite{li2024attention} is denoted as \textbf{CSID}, while the one by \cite{li2023multi} is labeled as \textbf{MCMF} in our results. We also examined the performance of the Transformer-based model \textbf{MulT} \cite{tsai2019multimodal} across both datasets.

\subsection{Effectiveness Comparison}

\begin{table*}
  \centering
  \resizebox{\textwidth}{!}{
  \begin{tabular}{|l|cccc|cccc|cccc|c|}
    \hline
    \multirow{2}{*}{\textbf{Modality}}  &
    \multicolumn{4}{|c|}{\textbf{Non Hate Videos}} &
    \multicolumn{4}{|c|}{\textbf{Implicit Hate Videos}} &
    \multicolumn{4}{|c|}{\textbf{Explicit Hate Videos}} &
    \textbf{Overall} \\
    & Acc & F1 & Prec & Rec & Acc  & F1 & Prec & Rec & Acc  & F1 & Prec & Rec &Macro-F1 \\
    \hline
    Text &
    0.8905 & 0.8424 & 0.7991 & 0.8905 & 0.2264 & 0.3038 & 0.4615 & 0.2264 & 0.7447 & 0.6393 & 0.5600 & 0.7447 & 0.5951 \\
    Image &
    0.8607 & 0.8317 & 0.8046 & 0.8607 & 0.2453 & 0.3077 & 0.4127 & 0.2453 & 0.7234 & 0.6267 & 0.5587 & 0.7234 & 0.5587 \\
    Audio &
    0.7811 & 0.7677 & 0.7548 & 0.7811 & 0.4151 & 0.4583 & 0.5116 & 0.4151 & 0.6064 & 0.5672 & 0.5327 & 0.6064 & 0.5977 \\
    \hline
    Text + Audio &
    0.8657 & 0.8208 & 0.7803 & 0.8657 & 0.3113 & 0.3750 & 0.4714 & 0.3113 & 0.6915 & 0.6435 & 0.6019 & 0.6915 & 0.6131 \\
    Audio + Image &
    0.8706 & 0.8413 & \textbf{0.8140} & 0.8706 & 0.2642 & 0.3394 & 0.4746 & 0.2642 & \textbf{0.7872} & \textbf{0.6697} & 0.5827 & \textbf{0.7872} & 0.6168 \\
    Text + Image &
    0.6816 & 0.6903 & 0.6995 & 0.6816 & 0.1604 & 0.2716 & \textbf{0.8500} & 0.1604 & 0.5532 & 0.3728 & 0.2811 & 0.5532 & 0.4448 \\
    \hline
    \textbf{Proposed Method} &
    \textbf{0.8955} & \textbf{0.8448} & 0.8000 & \textbf{0.8955} & \textbf{0.6698} & \textbf{0.6605} & 0.6513 & \textbf{0.6698} & 0.4894 & 0.5702 & \textbf{0.6866} & 0.4894 & \textbf{0.6918} \\
    \hline
  \end{tabular}
  }

  \caption{Impact of Different Modalities on Multiclass Classification on ImpliHateVid Dataset}
  \label{tab:impact of modalities}
\end{table*}

\begin{table*}
  \centering
  \resizebox{\textwidth}{!}{
  \begin{tabular}{|l|cccc|cccc|cccc|c|}
    \hline
    \multirow{2}{*}{\textbf{Features}}  &
    \multicolumn{4}{|c|}{\textbf{Non Hate Videos}} &
    \multicolumn{4}{|c|}{\textbf{Implicit Hate Videos}} &
    \multicolumn{4}{|c|}{\textbf{Explicit Hate Videos}} &
    \textbf{Overall} \\
    
    & Acc & F1 & Prec & Rec & Acc  & F1 & Prec & Rec & Acc  & F1 & Prec & Rec &Macro-F1 \\
    \hline
    w/o Emotions &
    0.8209 & 0.8312 & \textbf{0.8418} & 0.8209 & 0.6415  & 0.6507 & 0.6602 & 0.6415 & \textbf{0.5851} & 0.5609 & 0.5392 & \textbf{0.5851} & 0.6809\\
    w/o Captions & 
    0.8258 & 0.8217 & 0.8177 & 0.8258 & 0.6509 & 0.6448 & 0.6389 & 0.6509 & 0.5425 & 0.5542 & 0.5667 & 0.5425 & 0.6736\\
    w/o Sentiments & 
    0.8408 & 0.8325 & 0.8244 & 0.8408 & 0.6226 & 0.6438 & \textbf{0.6667} & 0.6226 & 0.5638 & 0.5548 & 0.5463 & 0.5638 & 0.6770\\
    \hline
    \textbf{Proposed Method} & 
    \textbf{0.8955} & \textbf{0.8448} & 0.8000 & \textbf{0.8955} & \textbf{0.6698} & \textbf{0.6605} & 0.6513 & \textbf{0.6698} & 0.4894 & \textbf{0.5702} & \textbf{0.6866} & 0.4894 & \textbf{0.6918}\\
    \hline
  \end{tabular}
  }
  \caption{Impact of Different Features on Multiclass Classification on ImpliHateVid Dataset}
  \label{tab:impact of features}
\end{table*}

\subsubsection{Binary Classification}
 
\autoref{tab:performance-comparison} highlights the performance improvements of our proposed multimodal method over the best performing approaches in each category on the ImpliHateVid and HateMM datasets. On the ImpliHateVid dataset, our method achieves an F1-score of 87.73\%, which is approximately 10.5\%  higher than the best unimodal method, Wav2Vec2, which attains an F1-score of 77.24\%. When compared to the leading multimodal baseline, MulT, with an F1-score of 83.52\%, our approach shows a 4.21\% improvement. 

On the HateMM dataset, the differences are even more pronounced. Our method reaches an F1-score of 97.58\%, outperforming the best text-based model, BERT, which records an F1-score of 66.40\%, by roughly 31.18\%. Similarly, against the top multimodal method in this category, CSID, with an F1-score of 71.40\%, our proposed approach gains about 26.18\%. 
These substantial gains in F1-score clearly demonstrate the effectiveness of our multimodal framework in leveraging complementary cues from text, image, and audio modalities, thereby significantly enhancing implicit hate speech detection across different datasets.

\subsubsection{Multiclass Classification}

\autoref{tab:multiclass-performance} presents the multiclass classification results across non-hate, implicit hate, and explicit hate video categories. Our proposed method achieves an overall macro-F1 of 69.18\%, which represents a significant improvement over the best unimodal approaches and existing multimodal baselines. For instance, among unimodal models, the text-based BERT attains a macro-F1 of 59.07\%, while the image-based ViT and audio-based Wav2Vec2 achieve 58.83\% and 57.16\%, respectively. This indicates that our model improves the macro-F1 score by roughly 10.1\% points over the best unimodal method. GPT and Llama fail to identify instances of implicit hate accurately due to noisy transcriptions. When compared to multimodal baselines, our approach also demonstrates clear superiority. The strongest competing multimodal method, MulT, records a macro-F1 of 66.27\%; hence, our proposed model exhibits an absolute improvement of approximately 2.91\%. Additionally, our model consistently outperforms other multimodal methods such as DeepCNN, CMHFM, CSID, and MCMF across individual categories such as non-hate, implicit hate, and explicit hate, highlighting its balanced performance. These results underscore the effectiveness of our two-stage contrastive learning framework in integrating complementary textual, visual, and audio cues, thereby establishing a new state-of-the-art for multiclass hate speech detection in videos.

\subsection{Ablation Analysis}

\subsubsection{Impact of Different Modalities}
\begin{figure}
    \centering
    \includegraphics[width=\columnwidth]{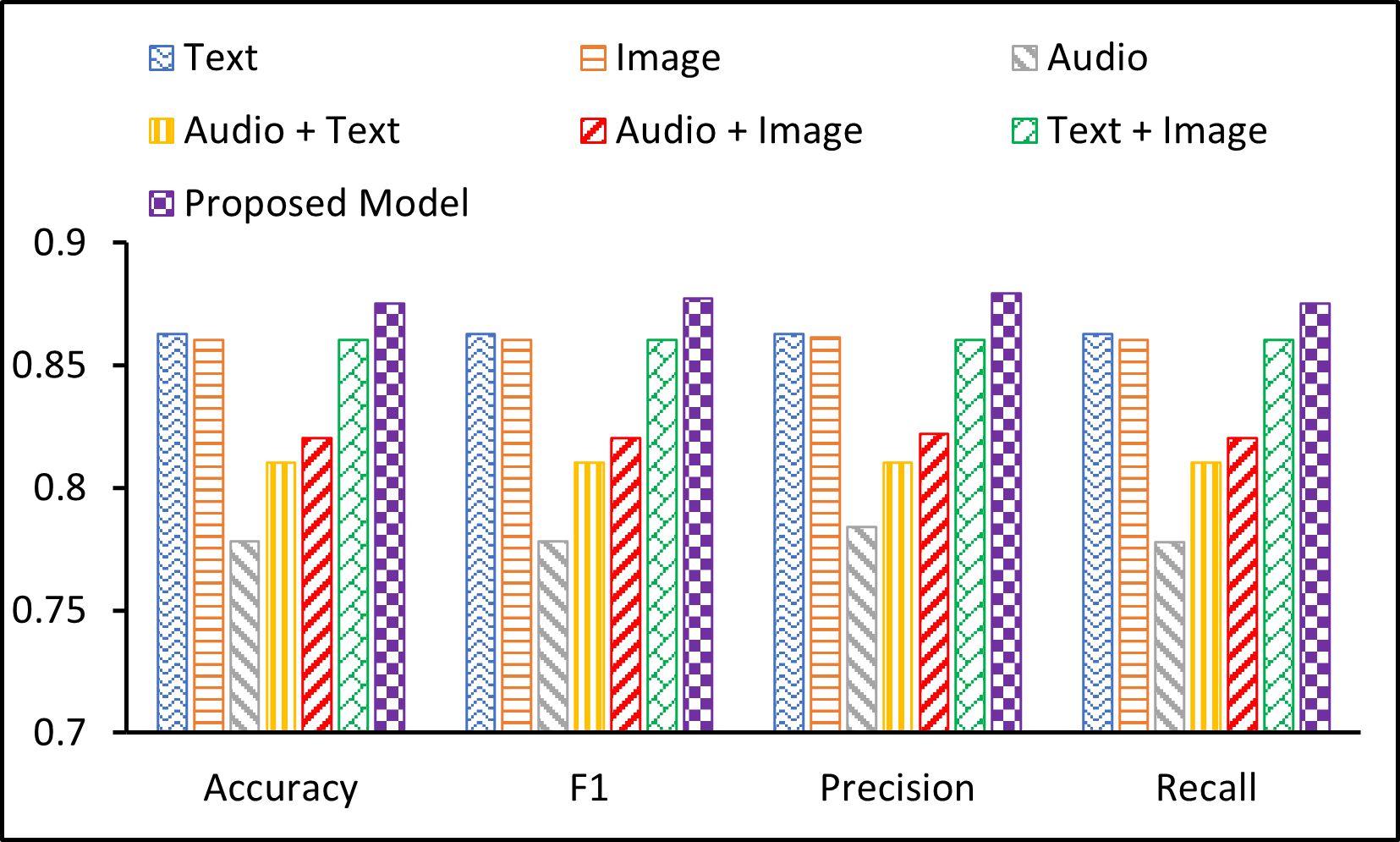}
    \caption{Impact of Different Modality Combinations on Binary Classification on ImpliHateVid}
    \label{fig:modality-impact}
\end{figure}

We compared the performance of our model, considering all combinations of the three modalities on the ImpliHateVid dataset. The results of binary classification can be seen in  \autoref{fig:modality-impact} and those of three-class classification have been highlighted in \autoref{tab:impact of modalities}.

Examining the table for unimodal results, the text-only model achieves strong performance for non-hate videos with an F1 of 84.24\% but struggles with implicit hate content with F1 of 30.38\%. Similarly, the image-only and audio-only models show moderate performance, with audio achieving a relatively higher F1 of 45.83\% on implicit hate videos compared to text and image modalities.

 The text + audio combination boosts performance on implicit hate videos with an of F1 of 37.50\% compared to using text alone. The audio + image configuration yields a notable improvement for explicit hate videos with an F1 of 66.97\%, highlighting the benefit of integrating complementary visual and auditory cues. Conversely, the text + image combination, despite showing promise in precision, falls short in overall effectiveness, as indicated by a lower macro-F1 of 44.48\%.

Our proposed method, which integrates textual, visual, and audio features simultaneously, outperforms all the ablated configurations. It achieves an overall accuracy of 89.55\%, an F1-score of 84.48\% for non-hate videos, and significantly higher performance for implicit hate with an F1 of 66.05\% and explicit hate videos with an F1 of 57.02\%. The overall macro-F1 of 69.18\% clearly demonstrates the advantage of fully leveraging multimodal information. \autoref{fig:modality-impact} further illustrates this performance gain, where the proposed model outperforms all other configurations in binary classification. These results emphasize that combining all three modalities leads to a more robust and balanced classification outcome across all categories.

\subsubsection{Impact of Emotion, Caption, and Sentiment Features}

\begin{figure}
    \centering
    \includegraphics[width=\columnwidth]{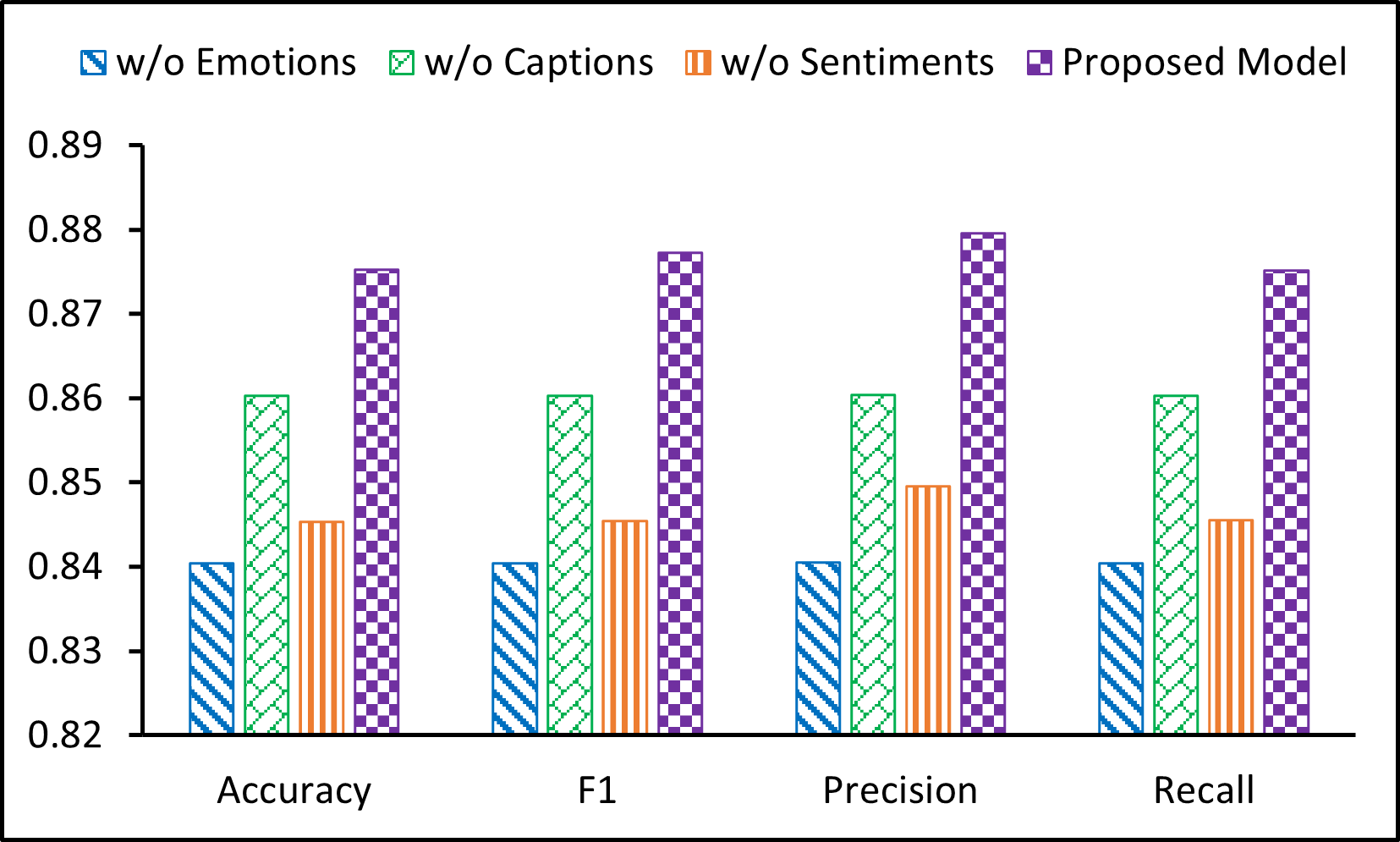}
    \caption{Impact of Different Features on Binary Classification on ImpliHateVid}
    \label{fig:feature-impact}
\end{figure}

\autoref{tab:impact of features} summarizes the impact of omitting specific features on the three-class classification performance. For each ablation, removing emotions, captions, or sentiments, the table reports accuracy, F1-score, precision, and recall for each video category, along with the overall macro-F1 score and overall accuracy. Notably, the full proposed model, which uses all features, achieves a balanced performance with an overall accuracy of 71.32\% and a macro-F1 of 66.29\%, outperforming the ablated versions. The binary classification results are shown in \autoref{fig:feature-impact}.

\section{Conclusion}

This work makes two significant contributions. First, we introduce ImpliHateVid, a novel dataset specifically curated for implicit hate speech detection in videos. Comprising 2,009 videos, ImpliHateVid represents one of the first large-scale benchmarks for video-based implicit hate detection. This dataset not only fills a critical gap in the literature but also provides a valuable resource for advancing research in multimodal hate speech analysis. Second, we propose a two-stage contrastive learning framework that effectively integrates textual, visual, and audio modalities. In the first stage, dedicated encoders extract robust, modality-specific features. These features are then aligned into a unified embedding space via cross-modal encoders in the second stage, using a supervised contrastive loss that clusters similar samples and separates dissimilar ones. Additionally, our model is enhanced by incorporating sentiment, emotion, and image caption features, which capture subtle cues associated with hate speech. Extensive experiments on ImpliHateVid and the HateMM dataset demonstrate that our approach significantly outperforms all the baselines. These findings underscore the effectiveness of leveraging cross-modal information to capture the full context of hateful content in videos. Future work will focus on extending the framework to incorporate multilingualism and exploring real-time detection applications. Overall, our contributions pave the way for more accurate systems in combating online hate in videos.

\section{Limitations}

Despite the promising results, our approach has several limitations. First, the performance of the multimodal framework depends heavily on the quality and alignment of data across modalities. Noisy or misaligned text, image, or audio inputs can adversely affect the joint embedding space and, consequently, the overall classification accuracy. Second, the reliance on pre-trained encoders means that any shortcomings in these models, such as domain mismatch or insufficient representation of hate speech nuances, can limit the effectiveness of our system. 

Third, the supervised contrastive loss, while effective, is sensitive to hyperparameter settings such as the temperature parameter and the strategy for selecting positive and negative pairs. Improper tuning can lead to suboptimal clustering of similar samples and separation of dissimilar ones. 
Future research should focus on developing more robust and adaptive models and expanding the dataset to cover a broader range of hate speech phenomena.

\bibliography{acl_text}

\end{document}